\documentclass[conference]{IEEEtran}
\IEEEoverridecommandlockouts

\usepackage{graphicx}
\usepackage{amsmath}
\usepackage{amssymb}
\usepackage{cite}

\usepackage{tikz}

\title{Weakly Supervised Learning on Large Graphs}

\author{
    \IEEEauthorblockN{Aditya Prakash}
    \IEEEauthorblockA{
        Indian Institute of Technology Bombay \\
         210260004@iitb.ac.in
    }
}

\begin{document}

\maketitle

\begin{abstract}
Graph classification plays a pivotal role in various domains, including pathology, where images can be represented as graphs.In this domain, images can be
represented as graphs, where nodes might represent individual nuclei, and edges
capture the spatial or functional relationships between them. Often, the overall
label of the graph, such as a cancer type or disease state, is determined by patterns
within smaller, localized regions of the image.  This work introduces a weakly-supervised graph classification framework leveraging two subgraph extraction techniques: (1) Sliding-window approach (2) BFS-based approach. Subgraphs are processed using a Graph Attention Network (GAT), which employs attention mechanisms to identify the most informative subgraphs for classification. Weak supervision is achieved by propagating graph-level labels to subgraphs, eliminating the need for detailed subgraph annotations. 
\end{abstract}

\section{Introduction}
Graph classification is a critical task in fields like cheminformatics, social networks, and medical diagnostics. In pathology, graphs represent images with nodes corresponding to cellular structures and edges capturing spatial relationships. Disease-relevant patterns are often localized in specific subregions of these graphs, making subgraph-based approaches essential.

Traditional graph classification methods require detailed annotations, which are expensive and time-consuming to obtain. To address this, we propose a weakly-supervised graph classification framework that uses graph-level labels to learn subgraph-level patterns.

Our approach introduces two subgraph extraction techniques:
\begin{itemize}
    \item \textbf{BFS-based extraction:} Captures connected and meaningful subgraphs using Breadth-First Search.
    \item \textbf{Sliding-window extraction:} Iteratively selects localized subgraphs using a fixed window size.
\end{itemize}

These subgraphs are processed using a Graph Attention Network (GAT), which identifies and aggregates the most informative subgraphs using attention scores. The proposed method is evaluated on the D\&D and MSRC-21 datasets, achieving competitive accuracy and providing interpretable insights.

\section{Related Work}
Graph Neural Networks (GNNs) have emerged as powerful tools for learning representations on graph-structured data. They extend the capabilities of traditional neural networks to non-Euclidean domains by leveraging the relationships encoded in graph edges. Among the most prominent architectures is the Graph Convolutional Network (GCN) \cite{kipf2016semi}, which generalizes the convolution operation to graphs. GCNs update node features by aggregating information from neighboring nodes using a propagation rule, enabling the learning of rich graph-level representations. However, GCNs assume equal importance for all neighbors, which may not always align with the underlying graph structure.

To address this limitation, Graph Attention Networks (GATs) \cite{velickovic2017graph} introduce an attention mechanism that learns the relative importance of edges. By assigning attention weights to edges, GATs enable the model to prioritize more relevant neighbors during feature aggregation. This mechanism has proven particularly effective in capturing localized patterns in tasks like node classification and graph classification.

In the context of medical image analysis, particularly in pathology, several studies have employed GNNs for graph-based classification. For example, the work by \cite{sureka2020visualization} explores the use of GCNs for histopathology image classification by transforming image patches into graphs, where nodes represent image features and edges capture spatial relationships between them. Their method demonstrated the effectiveness of GCNs in learning from histopathology images and provided valuable visualizations of the learned graph structures. This approach aligns with our goal of applying graph-based models to pathology images, where localized, disease-specific regions need to be identified for accurate diagnosis.

Building on these foundational works, our approach integrates GNN-based models with two novel subgraph extraction techniques—BFS-based and sliding-window methods—to address weakly-supervised graph classification. These methods enable the identification of relevant subgraphs from pathology graphs, while the attention mechanism in GATs ensures the model focuses on the most informative subgraphs, achieving good performance without requiring detailed subgraph-level annotations.

\section{Methodology}

In this section, we describe the approach for weakly-supervised graph classification, focusing on two subgraph extraction methods—BFS-based and sliding-window-based. We then detail how these subgraphs are processed using a Graph Attention Network (GAT), and how weak supervision is applied through attention-based subgraph selection.

\subsection{Subgraph Extraction Techniques}

Effective subgraph extraction is essential for isolating meaningful regions of a graph that contribute to graph-level classification. The two subgraph extraction techniques used in this work are:

\subsubsection{BFS-based Subgraph Extraction}
The BFS-based subgraph extraction method captures subgraphs starting from random nodes and expanding using breadth-first search (BFS) until a depth limit is reached. 

\begin{itemize}
    \item Start from a randomly selected node in the graph.
    \item Traverse the graph using BFS up to a specified depth limit (e.g., 11).
    \item Collect all the nodes and edges encountered during the traversal.
    \item If the subgraph has fewer than a specified minimum number of nodes or edges, discard it.
    \item Retain the subgraphs that meet the minimum criteria for further processing.
\end{itemize}

\begin{figure}[h!]
    \centering
    \includegraphics[width=0.8\linewidth]{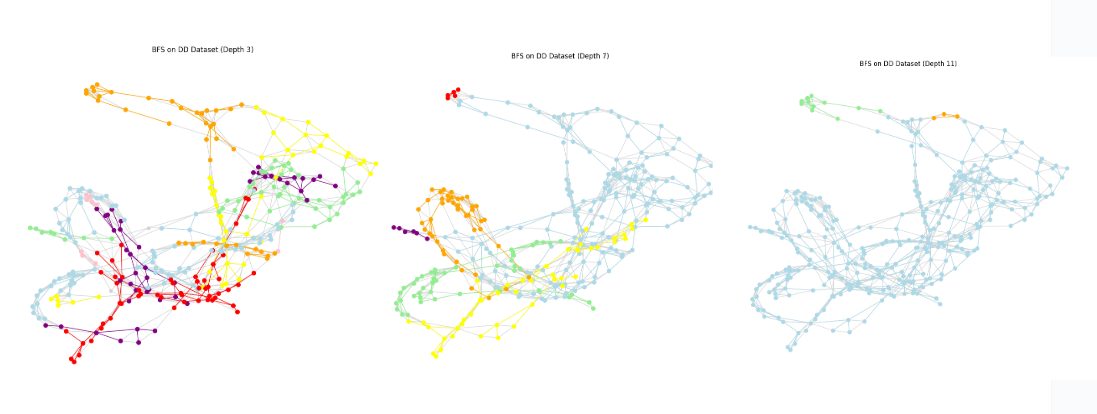}
   \caption{Different Depth Limit}
    
\end{figure}

\subsubsection{Sliding-Window Subgraph Extraction}
The sliding-window method extracts subgraphs by moving a fixed-size window across the nodes of the graph, creating overlapping subgraphs. This method is particularly useful for capturing localized graph regions and allows for greater flexibility in exploring different parts of the graph.

\begin{itemize}
    \item Define a window size and step size .
    \item For each position of the window, create a subgraph by selecting the nodes within the window.
    \item Filter the edges to include only those that connect nodes within the window.
    \item Retain the node features and graph-level label for each subgraph.
\end{itemize}

\begin{figure}[h!]
    \centering
    \includegraphics[width=0.8\linewidth]{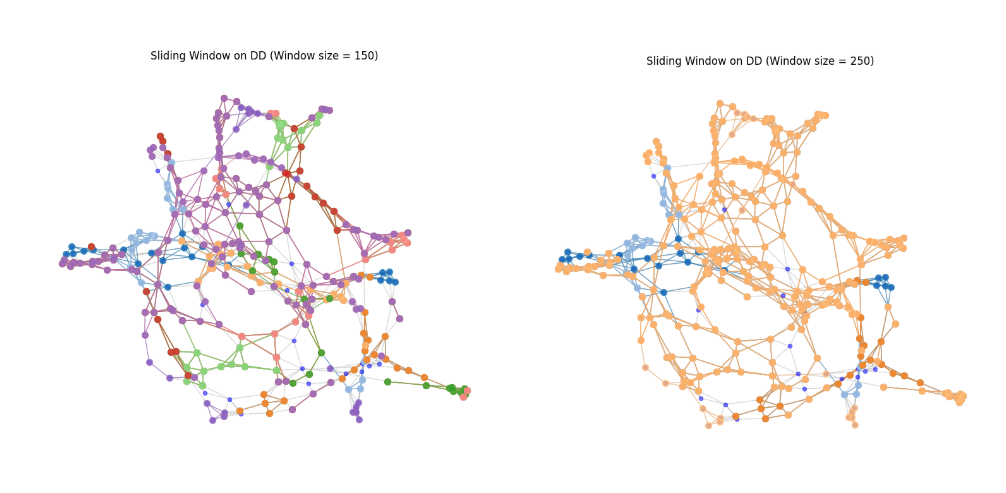}
    \caption{Different Window Sizes}

\end{figure}
\subsection{Model Architecture}

Once the subgraphs are extracted, they are processed using a \textbf{Graph Attention Network (GAT)}, a model designed to focus on the most relevant parts of a graph by assigning different attention weights to its edges and nodes. The GAT model applies attention mechanisms that allow it to prioritize important neighbors during feature aggregation, which is crucial for tasks like graph classification.

\subsubsection{Graph Attention Mechanism}

In a traditional Graph Convolutional Network (GCN), the node features are aggregated by averaging the features of neighboring nodes. However, this approach assumes equal importance for all neighbors. In contrast, GATs use an attention mechanism to dynamically assign different importance to each neighbor, allowing the model to focus on the most relevant nodes during aggregation.

The key idea behind the GAT is to compute an attention score \( \alpha_{ij} \) for each pair of nodes \( i \) and \( j \) in the graph, which indicates how much importance node \( j \) should have when aggregating node \( i \)’s features. The attention scores are computed using the following steps:
\begin{itemize}

\item{Attention Coefficients}

For each edge in the graph, the attention score \( \alpha_{ij} \) is computed as:
\[
\alpha_{ij} = \frac{\exp \left( \text{LeakyReLU} \left( \mathbf{a}^\top [ \mathbf{W} \mathbf{h}_i \parallel \mathbf{W} \mathbf{h}_j ] \right) \right)}{\sum_{k \in \mathcal{N}(i)} \exp \left( \text{LeakyReLU} \left( \mathbf{a}^\top [ \mathbf{W} \mathbf{h}_i \parallel \mathbf{W} \mathbf{h}_k ] \right) \right)}
\]
where:
- \( \mathbf{h}_i \) and \( \mathbf{h}_j \) are the feature vectors of nodes \( i \) and \( j \), respectively.
- \( \mathbf{W} \) is a learnable weight matrix that projects the node features to a higher-dimensional space.
- \( \mathbf{a} \) is a learnable attention vector.
- \( \parallel \) denotes concatenation.
- The denominator is the normalization term to ensure that the attention scores sum to 1 over all neighbors of node \( i \).

\item{Feature Aggregation}

Once the attention scores are computed, the node features are aggregated using a weighted sum of the neighbors' features, where the attention scores serve as the weights:
\[
\mathbf{h}_i' = \sigma \left( \sum_{j \in \mathcal{N}(i)} \alpha_{ij} \mathbf{W} \mathbf{h}_j \right)
\]
where:
- \( \mathbf{h}_i' \) is the updated feature for node \( i \).
- \( \mathcal{N}(i) \) denotes the neighbors of node \( i \).
- \( \sigma \) is a non-linear activation function, typically ReLU or LeakyReLU.

\end{itemize}
\subsubsection{Multi-Head Attention}

In practice, we use multi-head attention to stabilize learning and allow the model to jointly attend to information from different subspaces. The multi-head attention mechanism applies multiple attention heads independently, aggregates their outputs, and then concatenates them.

For \( K \) attention heads, the output of the multi-head attention for node \( i \) is computed as:
\[
\mathbf{h}_i' = \parallel_{k=1}^{K} \sigma \left( \sum_{j \in \mathcal{N}(i)} \alpha_{ij}^k \mathbf{W}^k \mathbf{h}_j \right)
\]
where:
- \( \alpha_{ij}^k \) and \( \mathbf{W}^k \) are the attention scores and weight matrices for the \( k \)-th attention head.

The multi-head attention helps the model to capture diverse patterns in the graph structure and node features.

\subsubsection{GAT Layer and Aggregation}

The GAT model consists of multiple layers, each of which applies the attention mechanism to update the node features. The first layer applies multi-head attention, and the second layer aggregates the node features using a single attention head.

After the attention mechanism processes the nodes, the resulting features are aggregated at the graph level using global mean pooling, which computes the mean of the features from all nodes in the graph:
\[
\mathbf{h}_{\text{graph}} = \frac{1}{|V|} \sum_{i \in V} \mathbf{h}_i'
\]
where \( |V| \) is the number of nodes in the graph, and \( \mathbf{h}_i' \) is the feature vector of node \( i \) after the GAT layers.

\subsubsection{Final Classification}

The aggregated graph-level features are then passed through a fully connected layer to produce the final classification output:
\[
\hat{y} = \text{softmax} \left( \mathbf{W}_{\text{out}} \mathbf{h}_{\text{graph}} + \mathbf{b} \right)
\]
where:
- \( \mathbf{W}_{\text{out}} \) is the weight matrix for the final layer.
- \( \mathbf{b} \) is the bias term.
- \( \hat{y} \) is the predicted class for the graph.

\subsection{Weakly-Supervised Learning Setup}

In this work, the focus is on weakly-supervised learning, where only graph-level labels are available, and no subgraph-level annotations exist. To address this challenge, we leverage the attention mechanism in the GAT to guide the model towards the most informative subgraphs during training and inference.

\begin{itemize}
    \item For each extracted subgraph, compute attention scores using the GAT.
    \item These scores indicate the importance of each node and edge.
    \item The top-K subgraphs with the highest attention scores are selected.
    \item The predictions from these top-K subgraphs are aggregated (e.g., by averaging their outputs).
    \item The final classification decision is made by applying softmax to the aggregated predictions and selecting the class with the highest probability.
\end{itemize}

This attention-based subgraph selection allows the model to focus on the most relevant portions of the graph despite the lack of detailed subgraph-level labels.

\section{Experimental Setup}

In this section, we describe the experimental setup, including the datasets used for evaluation, as well as the training and evaluation processes.

\subsection{Dataset}

For the evaluation of our method, we use two widely used datasets in the graph classification domain: the D\&D dataset and the MSRC-21 dataset.

\subsubsection{D\&D Dataset}
The D\&D dataset consists of graphs representing protein structures, where each graph corresponds to a protein structure, and the nodes represent amino acids while the edges represent the spatial proximity between them. The task is to classify these graphs into one of two classes: enzyme or non-enzyme. 

\subsubsection{MSRC-21 Dataset}
The MSRC-21 dataset is a graph-based dataset for image segmentation tasks, where graphs represent image patches. Each graph consists of nodes representing superpixels, and edges are defined based on the similarity between neighboring superpixels. The task is to classify these graphs into 21 different categories, with a total of 1,500 graphs. 

\subsection{Training and Evaluation}

The model is trained using the Adam optimizer, with a learning rate of \( 0.01 \) and weight decay of \( 5 \times 10^{-4} \) to prevent overfitting. The model is trained for a total of 100 epochs. 

For each dataset, we perform a train-test split where 80\% of the graphs are used for training and 20\% are used for testing. 
We apply dropout with a rate of 0.6 after each GAT layer to prevent overfitting during training.

\begin{itemize}
    \item With the Sliding Window approach, the model is trained with different window sizes for subgraphs.
    \item With the BFS approach, the model is trained for different depth limit subgraphs.
\end{itemize}

\section{Results and Discussion}
\subsection{Sliding Window Appraoch}

The baseline accuracy by training and testing with the same architecture is 72\% on D\&D  and 88\% on the MSRC dataset.
\\
The accuracy with sliding window subgraphs and proposed weakly supervised Setup, as shown in Tables~\ref{table:dd-results} and~\ref{table:msrc-results}.

\begin{table}[h!]
\centering
\caption{Accuracy on D\&D Dataset}
\begin{tabular}{|c|c|c|}
\hline
\textbf{Subgraph Size (\%)} & \textbf{Nodes} & \textbf{Accuracy} \\
\hline
30\% & 85 & 0.6452 \\
40\% & 114 & 0.6782 \\
50\% & 142 & 0.6973 \\
60\% & 170 & 0.7333 \\
70\% & 199 & 0.7741 \\
80\% & 227 & 0.7923 \\
\hline
\end{tabular}
\label{table:dd-results}
\end{table}

\begin{figure}[h!]
    \centering
    \includegraphics[width=0.8\linewidth]{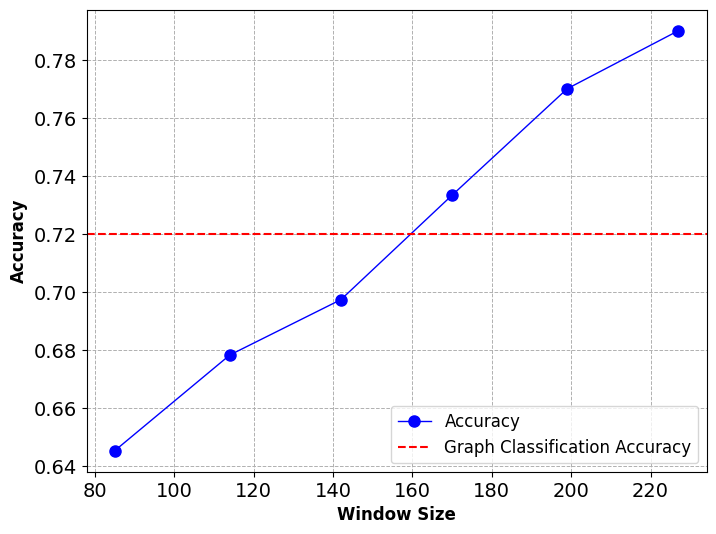}
    \caption{Accuracy vs Window Size in D\&D}
    
\end{figure}

\begin{table}[h!]
\centering
\caption{Accuracy on MSRC-21 Dataset}
\begin{tabular}{|c|c|c|}
\hline
\textbf{Subgraph Size (\%)} & \textbf{Nodes} & \textbf{Accuracy} \\
\hline
30\% & 23 & 0.8053 \\
40\% & 31 & 0.8584 \\
50\% & 39 & 0.8673 \\
60\% & 46 & 0.8584 \\
70\% & 54 & 0.8900 \\
80\% & 62 & 0.8667 \\
\hline
\end{tabular}
\label{table:msrc-results}
\end{table}

\begin{figure}[h!]
    \centering
    \includegraphics[width=0.8\linewidth]{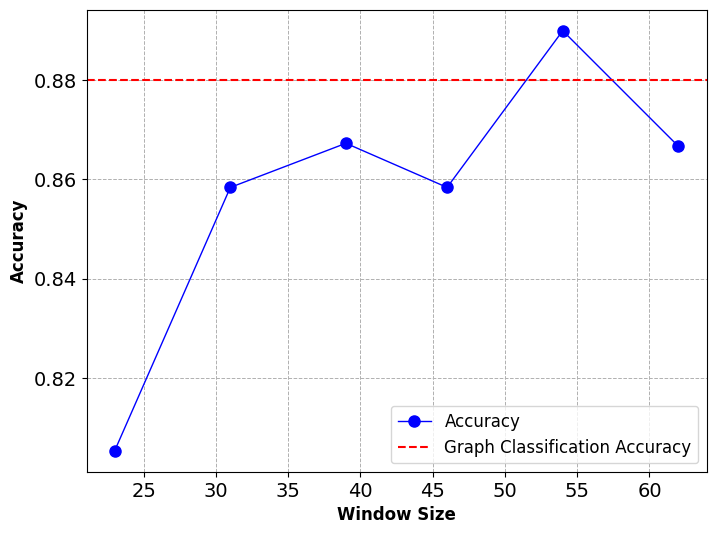}
    \caption{Accuracy vs Window Size in MSRC-21}
    
\end{figure}

\begin{figure}[h!]
    \centering
    \includegraphics[width=0.8\linewidth]{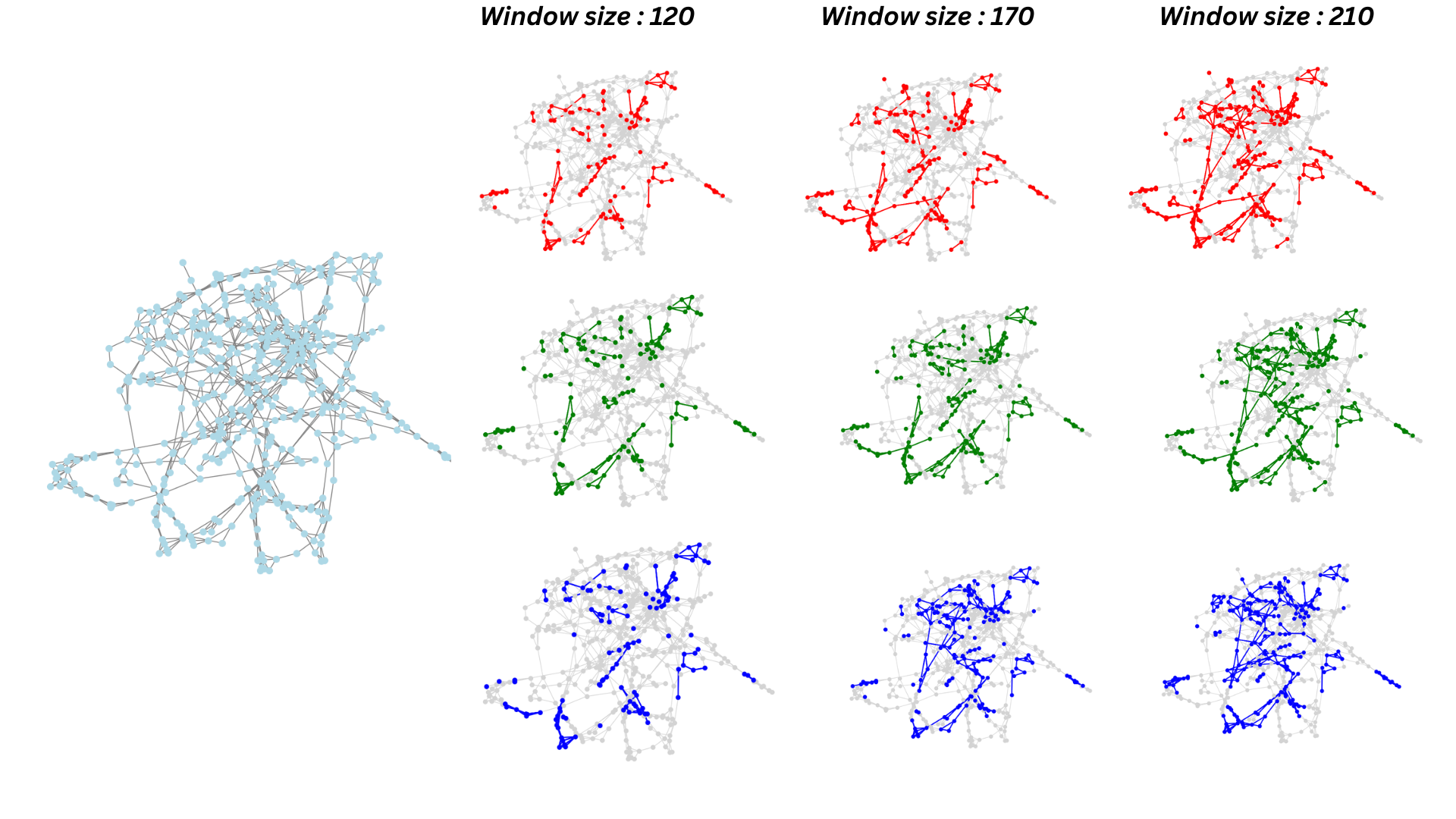}
    \caption{Top-k (k = 3) subgraphs according to attention scores (D\&D) }
    
\end{figure}

\begin{figure}[h!]
    \centering
    \includegraphics[width=0.8\linewidth]{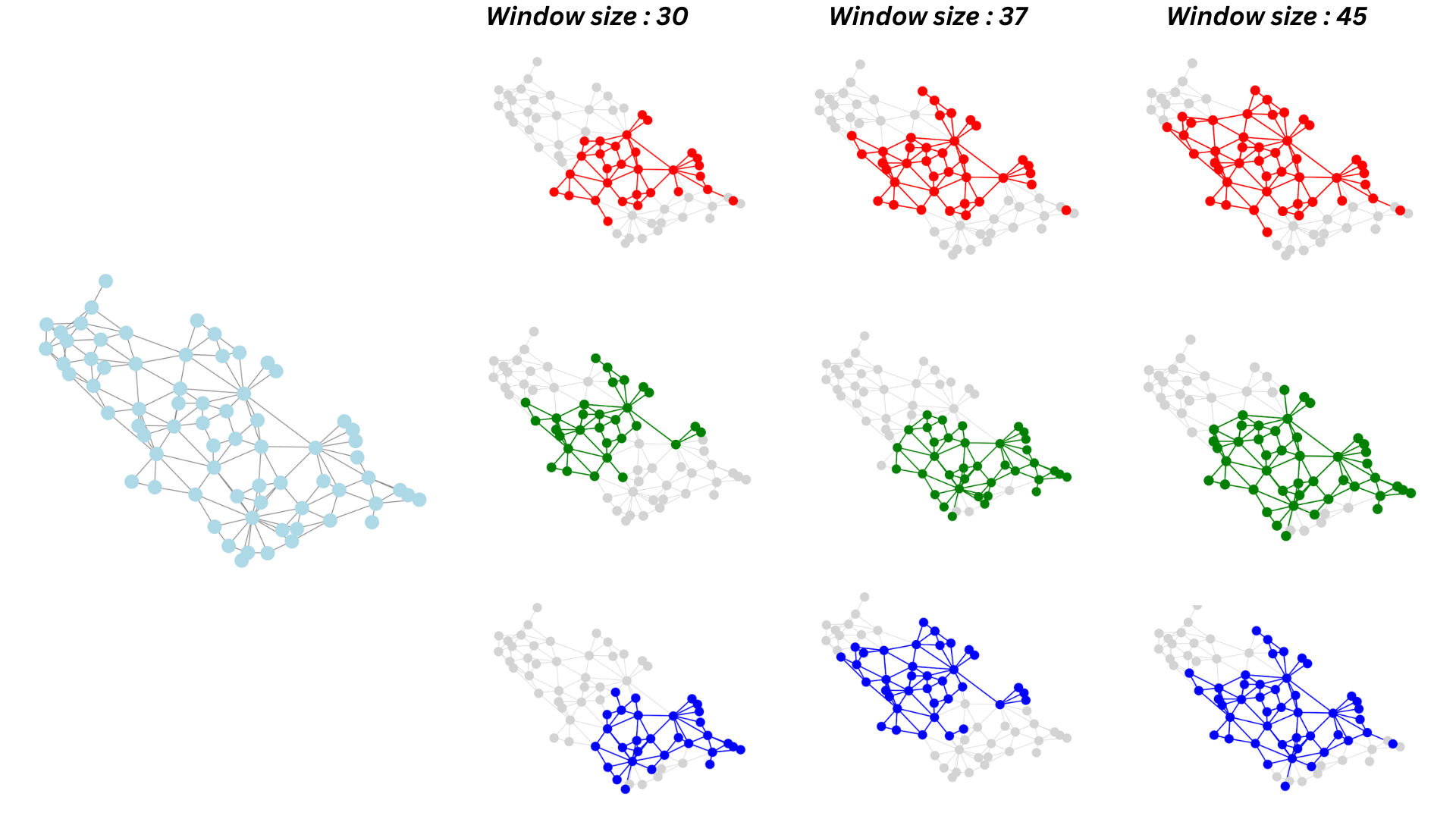}
    \caption{Top-k (k = 3) subgraphs according to attention scores (MSRC-21)}
    
\end{figure}

\subsection{BFS-based Appraoch}
The accuracy with BFS-based subgraphs and proposed weakly supervised Setup, as shown in Tables~\ref{table:dd-results} and~\ref{table:msrc-results}.

\begin{table}[h!]
\centering
\caption{Accuracy on D\&D Dataset}
\begin{tabular}{|c|c|c|}
\hline
\textbf{Depth Limit}  & \textbf{Accuracy} \\
\hline
3  & 0.5724 \\
5  & 0.5923 \\
7  & 0.6012 \\
9  & 0.6123 \\
11  & 0.6286 \\

\hline
\end{tabular}
\label{table:dd-results}
\end{table}

\begin{figure}[h!]
    \centering
    \includegraphics[width=0.8\linewidth]{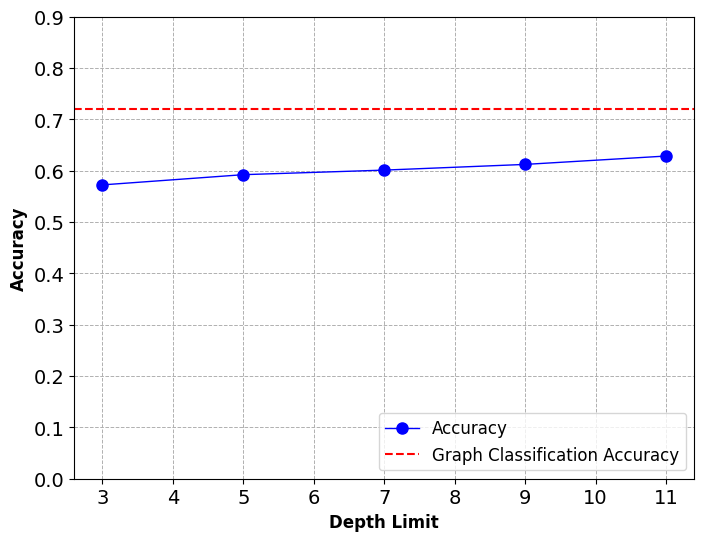}
    \caption{Accuracy vs Depth Limit in D\&D}
    
\end{figure}

\begin{table}[h!]
\centering
\caption{Accuracy on MSRC-21 Dataset}
\begin{tabular}{|c|c|c|}
\hline
\textbf{Depth Limit}  & \textbf{Accuracy} \\
\hline
2  & 0.8142 \\ 
4  & 0.7699 \\
6  & 0.7965 \\
8  & 0.8673 \\

\hline
\end{tabular}
\label{table:msrc-results}
\end{table}
\begin{figure}[h!]
    \centering
    \includegraphics[width=0.8\linewidth]{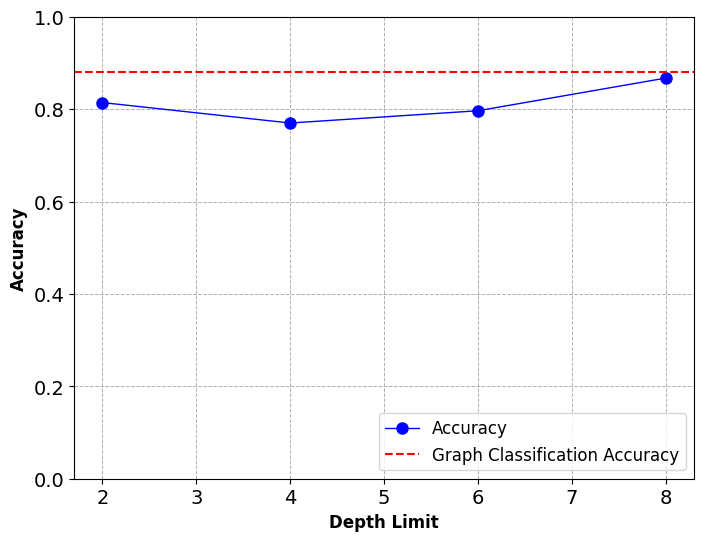}
    \caption{Accuracy vs Depth Limit in MSRC-21}
    
\end{figure}

    

    

\section{Conclusion and Future Work}

The report introduces a weakly-supervised graph classification framework utilizing BFS-based and sliding-window subgraph extraction techniques. The evaluation on D\&D and MSRC-21 datasets demonstrates competitive accuracy, particularly in datasets with higher node counts. The Graph Attention Network effectively identifies and prioritizes informative subgraphs, addressing the challenge of weak supervision. However, limitations persist, including instability in smaller datasets like MSRC and reliance on pre-defined hyperparameters.

Additionally, the current approach lacks qualitative insights into the extracted subgraphs, making it challenging to assess their relevance or interpretability. Applying this method to pathology image graphs presents an exciting direction for exploration. This would not only help assess the practical utility of the proposed approach but also provide insights into its ability to identify biologically or clinically meaningful subgraph patterns, thereby bridging the gap between model predictions and domain-specific knowledge.
\section*{Acknowledgments}
This work was conducted under the supervision of Prof. Amit Sethi at IIT Bombay.

\bibliographystyle{IEEEtran}
\bibliography{references}

\end{document}